\documentclass[10pt,twocolumn,letterpaper]{article}

\usepackage{times}
\usepackage{epsfig}
\usepackage{graphicx}
\usepackage{amsmath}
\usepackage{amssymb}
\usepackage[utf8]{inputenc} 
\usepackage[T1]{fontenc}    
\usepackage{url}            
\usepackage{booktabs}       
\usepackage{amsfonts}       
\usepackage{nicefrac}  
\usepackage{microtype} 
\usepackage{bm}
\usepackage{bbm}
\usepackage{times}
\usepackage{authblk}
\usepackage{epsfig}
\usepackage{algorithm}
\usepackage{algpseudocode}
\usepackage{xcolor,colortbl}
\usepackage{graphicx}
\usepackage{color}
\usepackage{stmaryrd}

\usepackage[pagebackref=true,breaklinks=true,letterpaper=true,colorlinks,bookmarks=false]{hyperref}

\begin{document}

\title{DeCaFA: Deep Convolutional Cascade for Face Alignment In The Wild}

\author[1, 2]{Arnaud Dapogny}
\author[2, 3]{Kevin Bailly}
\author[1]{Matthieu Cord}

\affil[1]{\small LIP6, Sorbonne Universit\'e, 4 Place Jussieu, Paris, France}
\affil[2]{\small Datakalab, 114 Boulevard Malesherbes, 75017 Paris}
\affil[3]{\small ISIR, Sorbonne Universit\'e, 4 Place Jussieu, Paris, France}

\date{}
\maketitle

\begin{abstract}
	Face Alignment is an active computer vision domain, that consists in localizing a number of facial landmarks that vary across datasets. State-of-the-art face alignment methods either consist in end-to-end regression, or in refining the shape in a cascaded manner, starting from an initial guess. In this paper, we introduce DeCaFA, an end-to-end deep convolutional cascade architecture for face alignment. DeCaFA uses fully-convolutional stages to keep full spatial resolution throughout the cascade. Between each cascade stage, DeCaFA uses multiple chained transfer layers with spatial softmax to produce landmark-wise attention maps for each of several landmark alignment tasks. Weighted intermediate supervision, as well as efficient feature fusion between the stages allow to learn to progressively refine the attention maps in an end-to-end manner. We show experimentally that DeCaFA significantly outperforms existing approaches on 300W, CelebA and WFLW databases. In addition, we show that DeCaFA can learn fine alignment with reasonable accuracy from very few images using coarsely annotated data.
\end{abstract}

\section{Introduction}

\begin{figure}[h!]
	\centering
	\includegraphics[width=\linewidth]{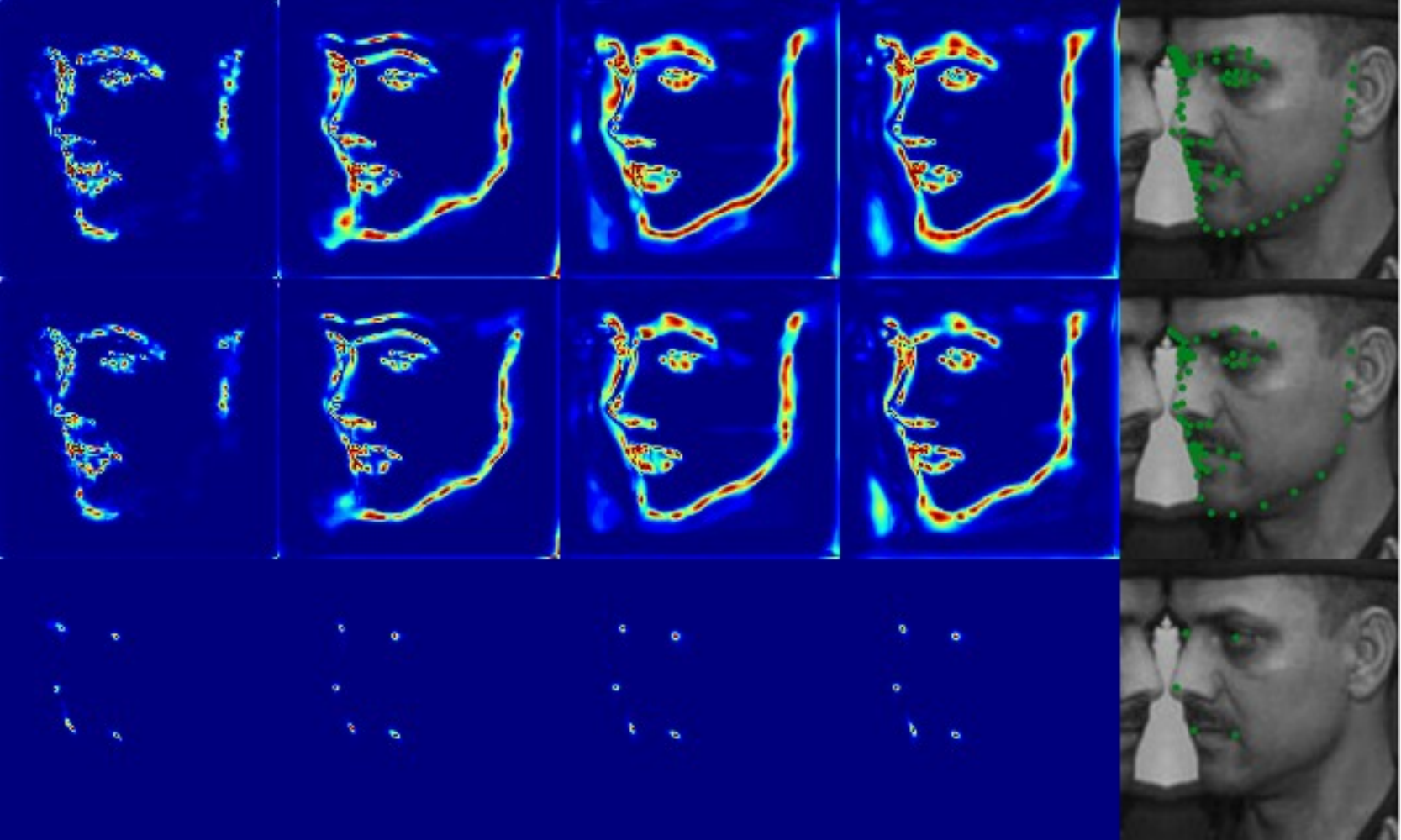}
	\caption{DeCaFA estimates landmark-wise attention maps at several stages of its architecture (horizontally: stages 1 to 4). By chaining transfer layers, it can integrate heterogeneous data (Vertically: attention maps and predictions for 98, 68 and 5-landmarks.}
	\label{main}
\end{figure}

Face alignment consists in localizing landmarks (e.g. lips and eyes corners, pupils, nose tip) on an face image. It is an important computer vision field, as it is an essential preprocess for face recognition \cite{taigman2014deepface}, tracking \cite{chrysos2018comprehensive}, expression analysis \cite{zhang2018bilateral}, and face synthesis \cite{thies2016face2face}.

Most recent face alignment approaches either belongs to cascaded regression methods, or to deep end-to-end regression methods. On the one's hand, cascaded regression consists in learning a sequence of updates, starting from an initial guess, to refine the landmark localization in a coarse-to-fine manner. This allows to robustly learn rigid transformations, such as translation and rotation, in the first cascade stages, while learning non-rigid deformation (e.g. due to facial expression or non-planar rotation) later on. 

On the other hand, many deep approaches aim at regressing the landmark position from the original image directly. However, because annotating several landmarks on a face image is a tedious task, data is rather scarce and the nature of the annotations usually vary a lot between the databases. Because of the scarcity of the data, end-to-end approaches usually rely on learning an intermediate representation, such as edges detection to drive the alignment process. However, these representations are usually ad hoc and do not guarantee to be optimal to address landmark localization tasks.

In this paper, we introduce a Deep convolutional Cascade for Face Alignment (DeCaFA). DeCaFA is composed of several stages that each produce landmark-wise attention maps, relatively to heterogeneous annotation markups. Figure \ref{main} shows attention maps extracted by the subsequent DeCaFA stages (horizontally) and for three different markups (vertically). It illustrates how these attention maps are refined through the successive stages, and how the different prediction tasks can benefit from each other. The contributions of this paper are tree-fold:

\begin{itemize}
	\item We introduce a fully-convolutional Deep Cascade for Face Alignment (DeCaFA) that unifies cascaded regression and end-to-end deep approaches, by using landmark-wise attention maps fused to extract local information around a current landmark estimate.
	
	\item We show that intermediate supervision with increasing weights helps DeCaFA to learn coarse attention maps in its early stages, that are refined in the later stages. Through chaining multiple transfer layers, DeCaFA integrates heterogeneous data annotated with different numbers of landmarks and model the intrinsic relationship between these tasks.
	
    \item We show experimentally that DeCaFA significantly outperforms existing approaches on multiple datasets, inluding the recent WFLW database. Additionally, we highlight how coarsely annotated data helps the network to learn fine landmark alignment even with very few annotated images.
\end{itemize}

\section{Related work}

Popular examples of cascaded regression methods include SDM \cite{Xiong2013}: in their pioneering work, Xiong \textit{et al} show that using simple linear regressors upon SIFT features in a cascaded manner already provides satisfying alignment results. LBF \cite{Ren2014} is a refinement that employs randomized decision trees to dramatically speed up feature extraction. DAN \cite{kowalski2017deep} uses deep networks to learn each cascade stage. However, one downside of these approaches is that the update regressors are not learned jointly in a end-to-end fashion, thus there is no guarantee that the learned feature point alignment sequences might be optimal. MDM \cite{Trigeorgis2016} improves the feature extraction process by sharing the convolutional layer among all steps of the cascade that are performed through a recurrent neural network. This results in memory footprint reduction as well as better representation learning and a more optimized landmark trajectory throughout the cascade.

TCDCN \cite{zhang2016learning} was perhaps the first end-to-end framework that could compete with cascaded regression approaches. It relies on supervised pretraining on a wide database of facial attributes. More recently, PCD-CNN \cite{kumar2018disentangling} uses head pose information to drive the training process. CPM+SBR \cite{dong2018supervision} employs landmark registration to regularize training. SAN \cite{dong2018style} uses adversarial networks to convert images from different styles to an aggregated style, upon which regression is performed. This aggregated style space thus serve as an intermediate representation that is more convenient for training. In \cite{wu2018look} the authors propose to use edge map estimation as an intermediate representation to drive the landmark prediction task, as well as to provide a unified representation when images are annotated in terms of different markups, that correspond to different alignment tasks. Finally, DSRN \cite{miao2018direct} relies on Fourier Embedding and low-rank learning to produce such representation. However, the use of such intermediate representation is usually ad hoc and it is hard to know which one would be all-around better for face alignment. Recently, AAN \cite{yue2018attentional} proposes to use intermediate feature maps as attentional masks to select relevant spatial regions. It also uses intermediate supervision to constrain those maps to correspond to attention maps relatively to landmark localization. However, there is no guarantee that the network will learn to align landmarks in a cascaded, coarse-to-fine manner.

Furthermore, annotating images in term of several face landmarks is a time-consuming task. As a result, data is rather scarce and annotated in terms of varying number of landmarks. For instance, 300W database \cite{Sagonas2015} contains approximately 3000 images labelled with 68 landmarks for train, whereas WFLW database \cite{wu2018look} contains 7500 images with 98 landmarks. Thus, one can wonder if we can use all those images within the same framework to learn more robust landmark predictions. In \cite{wu2017leveraging} the authors address this problem by using a classical multi-task formulation. However, this essentially ignores the intrinsic relationship between the structure of different landmark alignment tasks. Likewise, if we can predict the position of 68 landmarks, we can also easily deduce the position of landmarks for a coarser markup, such as eye/mouth corners and nose tip \cite{liu2015deep}. 

\section{DeCaFA overview}

\begin{figure*}[h!]
	\centering
	\includegraphics[width=\linewidth]{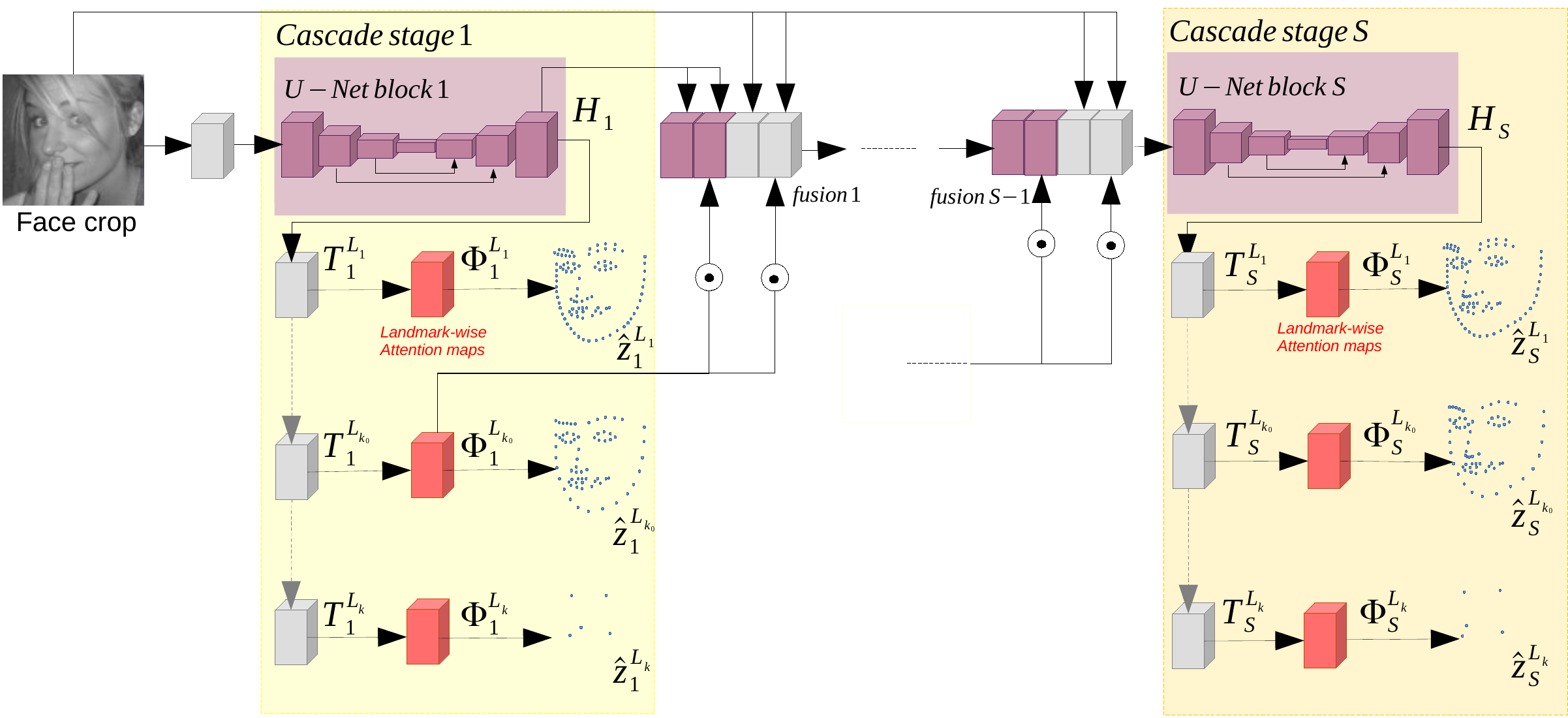}
	\caption{DeCaFA architecture overview. Several stages with fully-convolutional U-nets are stacked, multiple transfer layers are chained and intermediate supervision with increasing weights is applied to produce landmark estimates for heterogeneous alignment tasks. Landmark-wise attention maps are fused with the input image and the embeddings of the previous stage U-net to enable end-to-end cascaded alignment.}
	\label{overview}
\end{figure*}

In this Section, we introduce our Deep convolutional Cascade for Face Alignment (DeCaFA) model, as illustrated on Figure \ref{overview}. DeCaFA consists of $S$ stages, each of which contains a fully-convolutional U-net backbone that preserves the full spatial resolution, as well as an attention map generation sub-network. Section \ref{attmaps} shows how we derive landmark-wise attention maps for one landmark prediction task. Section \ref{chainattmaps} explains how several transfer layers can be chained to produce such attention maps, relatively to $K$ landmark prediction tasks. the input of the next stage is obtained by applying a feature fusion algorithm that involves the attention maps, as explained in Section \ref{fuuusion}. In Section \ref{leearning} we describe how DeCaFA is trained in an end-to-end manner with weighted intermediate supervision. Finally, in Section \ref{implemdetail} we provide implementation details to facilitate reproducibility of the results.

\subsection{Landmark-wise attention maps}\label{attmaps}

The U-net at stage $i$ takes an input $I_i$ and gives rise to an embedding $H_i$ with parameters ${\theta_i}$. In order to produce a suitable embedding from $H_i$ for predicting $L$ landmarks, we apply a $1 \times 1$ convolutional layer with $L$ filters with parameters ${\theta'_i}$. We denote the embeddings outputted by this transfer layer as ${T}^{L}_i$.
In order to highlight its dominant mode we apply a spatial softmax operator. Formally, for a pixel with coordinates $(x,y)$ and a landmark $l$:

\begin{equation}
\Phi^{L}_i(x,y,l)=\frac{\exp ({T}^{L}_i(x,y,l))}{\sum \limits_{x=1}^X \sum \limits_{y=1}^Y \exp ({T}^{L}_i (x,y,l))}
\end{equation}

An estimation $\hat z^{L}_i$ of the landmark coordinates can be obtained by computing the first order moments of $\Phi^{L}_i$:

\begin{equation}
\left \{
\begin{array}{l @{=} l}
\hat z_{i,x}^{L}(l)&\mathbb{E}_{x,y}[x\Phi^{L}_i(x,y,l)]\\
\hat z_{i,y}^{L}(l)&\mathbb{E}_{x,y}[y\Phi^{L}_i(x,y,l)]\\
\end{array}
\right.
\end{equation}

Where $\hat z_{i,x}^{L}$ and $\hat z_{i,y}^{L}$ are two vectors of size $L$ containing the $x$ and $y$ landmark coordinates $\hat z_i^{L}$. The soft-argmax operator is inspired by the work in \cite{luvizon2017human} in the frame of human pose estimation and provides differentiable landmark coordinates estimate from the attention map $\Phi^{L}_i$.

\subsection{Chaining landmark localization tasks}\label{chainattmaps}

As it will be explained in Section \ref{datasets}, existing datasets for face alignment usually have heterogeneous annotations and varying numbers of annotated landmarks. In order to deal with these heterogeneous annotations, we integrate $K$ tasks that consist in predicting various numbers of landmarks $L_1,...L_K$ with $\forall k_1,k_2$, $k_1 \leq k_2 \implies L_{k_1} > L_{k_2}$ (i.e. we chain the landmark-wise attention maps in an decreasing order of the number of landmarks to predict). To do so, we apply $K$ transfer layers $T^{L_{1}}_{i}, ..., T^{L_{K}}_{i}$ with parameters $\theta^{(1)}_i,...,\theta^{(K)}_i$, at it is depicted on Figure \ref{taskorder} (a). We have:

\begin{figure}[h!]
	\centering
	\includegraphics[width=\linewidth]{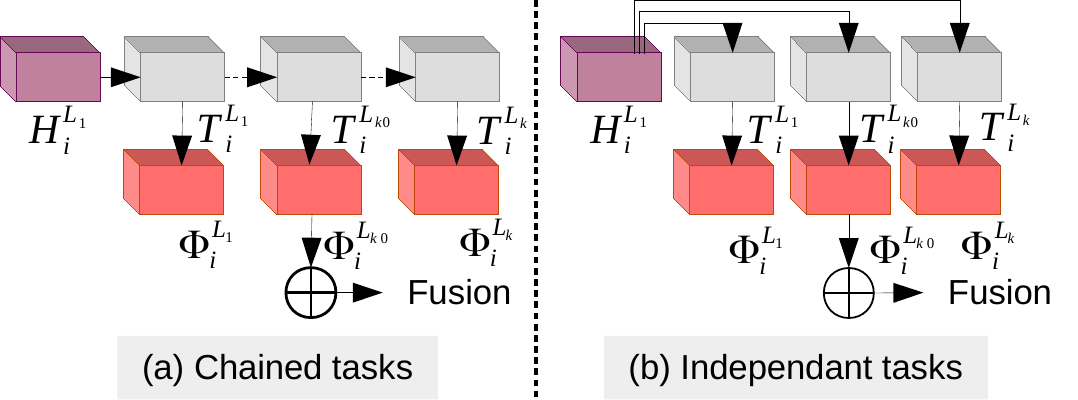}
	\caption{Chained (left) vs independant (right) task order.}
	\label{taskorder}
\end{figure}

\begin{equation}\label{fullmodel}
\left \{
\begin{array}{l @{=} l}
\hat z_{i,x}^{L_{k}}(l)&\mathbb{E}_{x,y}[x\Phi^{L_{k}}_{i}(x,y,l)] \quad \forall 1\leq k\leq K\\
\hat z_{i,y}^{L_{k}}(l)&\mathbb{E}_{x,y}[y\Phi^{L_{k}}_{i}(x,y,l)] \quad \forall 1\leq k\leq K\\
\end{array}
\right.
\end{equation}

The advantages of stacking the landmarks prediction pipelines in a descending order of the number of landmarks to be localized are two-fold:  First, from a semantic perspective, who can do more can do less, meaning that it shall be easier for the network to learn the sequence of transfer layers in that order (i.e. if we can precisely localize a 68-points markup it will be easy to also localize the nose tip, as well as mouth/eyes corners). Second, labelling images with large amounts of landmarks is a tedious task, thus generally the more annotated landmarks in a database, the less images we have at our disposal. Using such architecture ensures that the former (harder) tasks benefit from all the images annotated with the latter (easier) task. This can be seen as weakly supervised learning, where images labelled in terms of coarse markups can help to learn finer alignment tasks. Also note that as these $1 \times 1$ convolutional layers have very few parameters, thus a lot of gradient can be backpropagated down to the U-net backbone and benefit the $K$ prediction tasks. Finally, as illustrated on Figure \ref{taskorder}, we use attention maps $\Phi^{L_{k_0}}_i$ from markup $k_0$ to provide richer embeddings for the subsequent stages by applying feature fusion.

\subsection{Feature fusion}\label{fuuusion}

In a standard feedforward deep network with $S$ stacked stages, the $i+1^{th}$ stage takes an input $I_i=F_1$ that corresponds to the embeddings $H_i$ outputted by the previous stage (with the convention $I_0=I$ the original image). By contrast, in cascade-based approaches, each stage shall learn an update to bring the feature points closer to the ground truth localizations, by using information sampled around current feature point localizations. Within an end-to-end fully-convolutional deep network, an analogous statement would be that the $i+1^{th}$ stage shall use a local embedding $F_2$ that is calculated using information from the original image $I$ highlighted by landmark-wise attention maps $\Phi^{L_{k_0}}_i$. In our method, we aggregate these maps by summing all the landmark-wise attention maps $M_i=\bigoplus _{l=1}^L\Phi^{L_{k_0}}_i$. Thus, we can write the feature fusion model for the basic deep approach as:

\begin{equation}\label{f1}
F_{1}(I,H_i,M_i)=H_i
\end{equation}

and the cascade-like approach as:

\begin{equation}\label{f2}
F_{2}(I,H_i,M_i)=I \odot M_i
\end{equation}

Where $\odot$ denotes the Hadamard product. This fusion scheme between the input image and the mask only preserves local information, for which the values of $M_i$ are high. Alternatively, we can reinject the original image $I$ inside each stage so that it can use global information in case where the mask $M_i$ is not precise enough or contains localizations errors (as it is the case early in the cascade):

\begin{equation}\label{f3}
F_{3}(I,H_i,M_i)=I || (I \odot M_i)
\end{equation}

With $||$ the channel-wise concatenation operation. Furthermore we can also fuse the relevant parts (as highlighted by mask $M_i$) of the embedding $H_i$ of the previous stage U-net to provide the subsequent stages a richer, more semantically abstract information to estimate the landmarks coordinates:

\begin{equation}\label{f4}
F_{4}(I,H_i,M_i)=I || (I \odot M_i) || (H_i \odot M_i)
\end{equation}

Finally, we can aso use global information from not only the image $I$, but also from the embeddings $H_i$:

\begin{equation}\label{f5}
F_{5}(I,H_i,M_i)=I || (I \odot M_i) || H_i || (H_i \odot M_i)
\end{equation}

This fusion model is more efficient and is used in DeCaFA (Figure \ref{overview}), as it allows using both global and local information around the estimated landmarks so as to learn cascade-like alignment in an end-to-end fashion.

\subsection{Learning DeCaFA model}\label{leearning}

DeCaFA models can be trained end-to-end by optimizing the following loss function w.r.t. parameters of the U-nets $\theta_i$ and $\theta_i^{(1)},...,\theta_i^{(K)}$ for the transfer layers $T^{L_{1}}_{i}, ..., T^{L_{K}}_{i}$ respectively, $\forall 1\leq k\leq K$:

\begin{align}\label{lll}
\begin{split}
\mathcal{L}(\theta_1,\theta_1^{(1)},...,\theta_1^{(K)}&,...,\theta_S,\theta_S^{(1)},...,\theta_S^{(K)})=\\& \sum \limits_{k=1}^K \frac{1}{L_k}|\hat z_S^{L_k}-z^{L_k*}|
\end{split}
\end{align}

With $z^{L_k*}$ the ground truth landmark position for a $L_k$-landmarks markup. In practice, the summation in equation \eqref{lll} have less terms since usually each example is annotated with only one markup. With this configuration, however, if the whole network is deep enough, few gradient will ever pass through the firsts attention maps. Even worse, there is no guarantee that these feature maps will correspond to landmark-wise attention maps in the early stages, which is key to ensure cascade-like behavior of DeCaFA. To ensure this, we add a differentiable soft-argmax layer after each spatial softmax and a supervised cost at stage $i$:

\begin{align}\label{lossfun}
\begin{split}
\mathcal{L}(\theta_1,\theta_1^{(1)},...,\theta_1^{(K)}&,...,\theta_S,\theta_S^{(1)},...,\theta_S^{(K)})=\\& \sum \limits_{i=1}^S \lambda_i \sum \limits_{k=1}^K \frac{1}{L_k} |\hat z_i^{L_k}-z^{L_k*}|
\end{split}
\end{align}

In practice, we use a $\mathcal{L}_1$ loss function, as it has been shown to overfit less on very bad examples and lead to more precise results for face alignment. However, we need to make sure that the (relatively) shallow sub-networks does not overfit on these losses, which would result in very narrow heat maps with very localized dominant modes early in the cascade, and thus an overall lower accuracy. This is ensured by applying increasing $\lambda_i$ weights in \eqref{lossfun}.

\subsection{Implementation details}\label{implemdetail}

The DeCaFA models that will be investigated below use 1 to 4 stages that each contains 12 $3 \times 3$ convolutional layers with $64 \rightarrow 64 \rightarrow 128 \rightarrow 128 \rightarrow 256 \rightarrow 256$ channels for the downsampling portion, and vice-versa for the upsampling portion. The input images are resized to $128 \times 128$ grayscale images prior to being processed by the network. Each convolution is followed by a batch normalization layer with ReLU activation. In order to generate smooth feature maps we do not use transposed convolution but bilinear image upsampling followed with $3 \times 3$ convolutional layers. The whole architecture is trained using ADAM optimizer with a $5e^{-4}$ learning rate with momentum $0.9$ and learning rate annealing with power $0.9$. We apply $400000$ updates with batch size $8$ for each database, with alternating updates between the databases.

\begin{figure*}[h!]
	\centering
	\includegraphics[width=\linewidth]{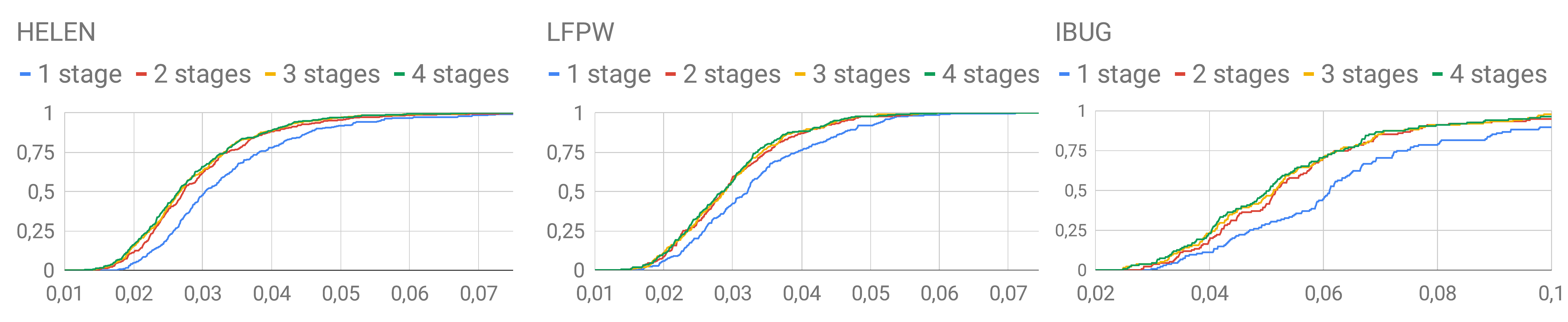}
	\caption{Comparison is terms of Cumulative error distribution (CED) curves on 300W of models with $S=$ 1,2,3 and 4 stages. As we stack cascade stages, the accuracy increases and saturates after the third/fourth stage.}
	\label{ced}
\end{figure*}

\begin{figure*}[h!]
	\centering
	\includegraphics[width=\linewidth]{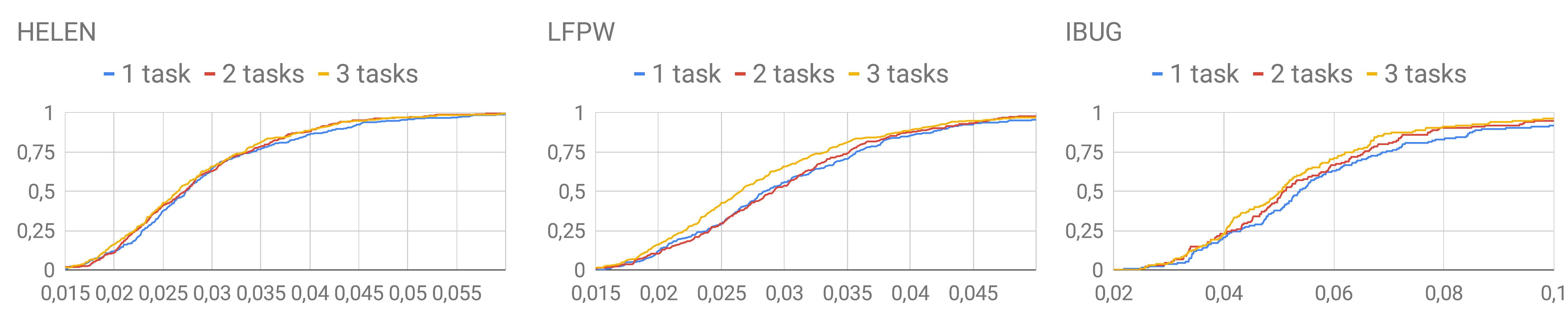}
	\caption{CED curves for models with $K=$ 1,2 and 3 landmark prediction tasks. Models trained with multiple alignment tasks are significantly better.}
	\label{ced2}
\end{figure*}

\begin{table*}[h!]
	\caption{Mean error (\%) comparison for multiple fusion, task ordering and intermediate supervision weighting schemes (lower is better). DeCaFA with $F_5$ fusion, chained tasks and increasing intermediate supervision weights performs better overall.}
	\label{ablation1}
	\begin{tabular}{l|l|l|r|r|r|r|r|r}
		\hline
		Fusion & task order& weights $\lambda_i$&300W-Full & 300W-Challenging & WFLW-All & WFLW-Pose & CelebA & Avg\\
		\hline
		$F_1$ &chained&$\uparrow$& \textbf{3.36}&5.27&4.71&8.3&2.53&4.83\\
		$F_2$  &chained&$\uparrow$& 3.45&5.45&4.83&8.78&2.70&5.04\\
		$F_3$ &chained &$\uparrow$& 3.43&5.38&4.76&8.39&\textbf{2.08}&4.81\\
		$F_4$ &chained&$\uparrow$& 3.40&5.31&4.65&8.25&2.41&4.80\\
		$F_5$ &chained&$\uparrow$& 3.39&\textbf{5.26}&\textbf{4.62}&\textbf{8.11}&2.10&\textbf{4.69}\\
		\hline
		$F_5$ &independant&$\uparrow$& 3.41  & 5.31 & 4.68 & 8.21 &2.16&4.75\\
		$F_5$ &chained&$=$& 3.41  & 5.33 & 4.84 & 8.77 &2.19&4.91\\
		$F_5$ &chained&$\downarrow$& 3.46  & 5.45 & 5.04 & 9.06 &2.23&5.05\\
		\hline
	\end{tabular}
\end{table*}

\section{Experiments}

\subsection{Datasets}\label{datasets}

The \textbf{300W} database, introduced in \cite{Sagonas2015}, is considered as the benchmark dataset for training and testing face alignment models, with moderate variations in head pose, facial expressions and illuminations. It consists in four databases: \textbf{LFPW} (811 images for train / 224 images for test), \textbf{HELEN} (2000 images for train / 330 images for test), \textbf{AFW} (337 images for train) and \textbf{IBUG} (135 images for test), for a total of 3148 images annotated with 68 landmarks for training the models. For comparison with state-of-the art methods, we refer to LFPW and HELEN test sets as the \textit{common} subset and I-BUG as the \textit{challenging} subset of 300W.

The \textbf{CelebA} database \cite{liu2015deep} is a large-scale face attribute database which contains $202599$ celebrity images coming from $10177$ identities, each annotated with $40$ binary attributes and the localization of $5$ landmarks (nose, left and right pupils, mouth corners). In our experiments, we train our models using the train partition that contains $162770$ images from $8k$ identities. The test partition contains $19962$ instances from $1k$ identities that are not seen in the train set.

The \textbf{Wider Facial Landmarks in the Wild or WFLW} database \cite{wu2018look} contains 10000 faces (7500 for training and 2500 for testing) with 98 annotated landmarks. This database also features rich attribute annotations in terms of occlusion, head pose, make-up, illumination, blur and expressions.

In what follows, we train our models using the train partition of 300W, WFLW and CelebA, and evaluate of the test partition of these datasets. As in \cite{Xiong2013,Ren2014,zhu2015face,zhang2016learning,miao2018direct,lv2017deep,xiao2016robust,honari2018improving} we measure the average point-to-point distance between feature points (ME), normalized by the inter-ocular distance (distance between outer eye corners on ground truth markup). As there is no consensus on how to measure the error we also report AUC and failure rates for a maximum error of $0.1$, as well as cumulative error distribution (CED) curves.

\subsection{Ablation study}

In this section, we validate the architecture and hyperparameters of our model: the number of stages $S$, the number of landmark prediction tasks $K$, the fusion and task ordering scheme as well as the intermediate supervision weights.

Figure \ref{ced} shows CED curves for models with $S=1,2,3$ and 4 cascade stages. The accuracy steadily increases as we add more stages, and saturates after the third on LFPW and HELEN, which is a well-known behavior of cascaded models \cite{Xiong2013,Ren2014}, showing that DeCaFA with weighted intermediate supervision indeed works as a cascade, by first providing coarse estimates and refining in the later stages. On IBUG, this difference is more conspicuous, thus there is for improvement by stacking more cascade stages.

Figure \ref{ced2} shows the interest of chaining multiple tasks, most notably on LFPW, that contains low-resolution images, and IBUG, which contains strong head pose variations as well as occlusions. Coarsely annotated data ($5$ landmarks) significantly helps the fine-grained landmark localization, as it is integrated a kind of weakly supervised scheme. This will be discussed more thoroughly in Section \ref{weeakly}.

Table \ref{ablation1} shows a comparison between multiple fusion, task ordering and intermediate supervision weighting schemes. We test our model on 300W (full and challenging), WFLW (All and challenging, \textit{i.e.} pose subset) as well as CelebA and report the average accuracy on those 5 subsets. First, reinjecting the whole input image ($F_3$ - Equation \eqref{f3} vs $F_2$ - Equation \eqref{f2}) significantly improves the accuracy on challenging data such as 300W-challenging or WFLW-pose, where the first cascade stages may commit errors. $F_4$ - Equation \eqref{f4} and $F_3$ fusion (cascaded models) using local+global information rivals the basic deep approach $F_1$ - Equation \eqref{f1}. Furthermore, $F_5$ - Equation \eqref{f5} fusion, which uses local and global cues is the best by a significant margin.

Furthermore, chaining the transfer layers (Figure \ref{taskorder}-a) is better than using independant transfer layers (Figure \ref{taskorder}-b): likewise, in such a case, the first transfer layer benefits from the gradients from the subsequents layer at train time. Last but not least, using increasing intermediate supervision weights in Equation \eqref{lossfun} (\textit{i.e.} $\lambda_1=1/8$, $\lambda_2=1/4$, $\lambda_3=1/2$, $\lambda_4=1$) is better than both using constant weights ( $\lambda_1=\lambda_2=\lambda_3=\lambda_4=1$) and decreasing weights ($\lambda_1=1$, $\lambda_2=1/2$, $\lambda_3=1/4$, $\lambda_4=1/8$), as it enables proper cascade-like training of the network, with the first stage outputting coarser attention maps that can be refined in the latter stages of the network.

\subsection{Comparisons with state-of-the-art methods}

\begin{table*}[ht]
	\caption{Comparison in terms of Mean error (lower is better), AUC (higher is better) as well as failure rate (lower is better), on WFLW.}
	\label{ablation4}
	\begin{tabular}{l|l|r|r|r|r|r|r|r}
		\hline
		metric&method&	all	&head pose&	expression&	illumination&	make-up	&occlusion&	blur\\
		\hline
		ME (\%)&CFSS \cite{zhu2015face} & 9.07 & 21.36 & 10.09 & 8.30 & 8.74 & 11.76 & 9.96 \\
		&DVLN \cite{wu2017leveraging}&10.84&46.93&11.15&7.31&11.65&16.30&13.71\\
		&LAB \cite{wu2018look}	&	5,27&	10,24&	5,51&	5,23&	5,15&	6,79&	6,32\\
		&Wing \cite{feng2018wing} & 5.11 & 8.75 & 5.36 & 4.93 & 5.41 & 6.37 & 5.81\\
		\hline
		&DeCaFA&	\textbf{4.62}&	\textbf{8.11}&	\textbf{4.65}&	\textbf{4.41}&	\textbf{4.63}&	\textbf{5.74}&	\textbf{5.38}\\
		\hline
		AUC@0.1&CFSS \cite{zhu2015face} & 0.366 & 0.063 & 0.316 & 0.385 & 0.369 & 0.269 & 0.303 \\
		&DVLN \cite{wu2017leveraging} &0.456&0.147&0.389&0.474&0.449&0.379&0.397\\
		&LAB \cite{wu2018look}	&	0.532&	0.235& 0.495&	0.543&	0.539&	0.449&	0.463\\
		&Wing \cite{feng2018wing} & 0.554 & \textbf{0.310} & 0.496 & 0.541 & 0.558 & \textbf{0.489} & 0.492\\
		\hline
		&DeCaFA&	\textbf{0.563}&	0.292&	\textbf{0.546}&	\textbf{0.579}&	\textbf{0.575}&	0.485&	\textbf{0.494}\\
		\hline
		FR@0.1(\%)&CFSS \cite{zhu2015face} & 20.56 & 66.26 & 23.25 & 17.34 & 21.84 & 32.88 & 23.67 \\
		&DVLN \cite{wu2017leveraging}&10.84&46.93&11.15&7.31&11.65&16.30&13.71\\
		&LAB \cite{wu2018look}	&	7.56&	28.83&	6.37&	6.73&	\textbf{7.77}&	13.72&	10,74\\
		&Wing \cite{feng2018wing} & 6.00 & 22.70 & 4.78 & 4.30 & 7.77 & 12.50 & 7.76\\
		\hline
		&DeCaFA&	\textbf{4.84}&	\textbf{21.4}&	\textbf{3.73}&	\textbf{3.22}&	\textbf{6.15}&	\textbf{9.26}&	\textbf{6.61}\\
		\hline
	\end{tabular}
\end{table*}

\begin{table*}[ht]
\centering
\begin{minipage}{0.5\textwidth}
\centering
\caption{Mean error (ME \%) comparison on 300W (lower is better).}
	\label{ablation2}
	\begin{tabular}{l|r|r|r}
		\hline
		Method & Common & Chall. & full \\
		\hline
		PCD-CNN \cite{kumar2018disentangling} & 3.67&7.62&4.44\\
		CPM+SBR \cite{dong2018supervision}&3.28&7.58&4.10\\
		SAN \cite{dong2018style} &3.34&6.60&3.98\\
		DAN \cite{kowalski2017deep} & 3.19 & 5.24&3.59\\
		LAB \cite{wu2018look} & 2.98  & 5.19  & 3.49  \\
		DAN-MENPO \cite{kowalski2017deep} & 3.09 & \textbf{4.88} & 3.44\\
		\hline
		DeCaFA & \textbf{2.93}  & 5.26  & \textbf{3.39}  \\
		\hline
	\end{tabular}
\label{labelrepclass}
\end{minipage}\hfill
\begin{minipage}{0.5\textwidth}
\centering
\caption{AUC and Failure rates (FR \%) for a maximum error of 0.1, and comparison with state-of-the-art approaches on 300W.}
	\label{ablation22}
	\begin{tabular}{l|r|r}
		\hline
		Method & AUC@0.1 & FR @0.1 (\%) \\
		\hline
		CFSS \cite{zhu2015face} &49.87&5.08\\
		Densereg+MDM \cite{alp2017densereg} & 52.19 & 3.67\\
		JMFA \cite{deng2017joint} &54.9&1.00\\
		JMFA-MENPO \cite{deng2017joint} &60.7&0.33\\
		LAB \cite{wu2018look} & 58.9  & 0.83    \\
		\hline
		DeCaFA & \textbf{0.661}  & \textbf{0.15}   \\
		\hline
	\end{tabular}
\end{minipage}\hfill
\end{table*}

Table \ref{ablation2} shows a comparison between DeCaFA and recent state-of-the-art approaches on 300W database. Our approach performs better than most existing approaches on the common subset, and performs very close to its best contenders on the challenging subset. Note that DeCaFA trained only on 300W trainset has a ME of $3.69\%$ and is already very competitive with recent approaches \cite{kumar2018disentangling,dong2018supervision,dong2018style,kowalski2017deep}, thanks to its end-to-end cascade architecture. DeCaFA is competitive with the best approaches, LAB \cite{wu2018look} and DAN-MENPO \cite{kowalski2017deep} as well as JMFA-MENPO \cite{deng2017joint}, which also use external data.

Table \ref{ablation4} shows a comparison between our method and LAB \cite{wu2018look} on WFLW database. As in \cite{wu2018look} we report the average point-to-point error on WFLW test partition, normalized by the outer eye corners. We also report the error on multiple test subsets containing variations in head pose, facial expressions, illumination, make-up as well as partial occlusions and occasional blur. DeCaFA performs better than LAB \cite{wu2018look} and Wing \cite{feng2018wing} by a significant margin on every subset. Also, note that DeCaFA trained solely on WFLW already as a ME of $5.01$ on the whole test set, which is still better that these two methods. Lastly, there is room for improvement on this benchmark as we do not excplicitely handle any of the factors of variation such as pose or occlusions.

\begin{table}[h!]
	\caption{Comparison with state-of-the-art approaches on CelebA database (lower is better).}
	\label{ablation3}
	\begin{tabular}{l|r}
		\hline
		Method & Mean error (\%) \\
		\hline
		SDM \cite{Xiong2013}   & 4.35  \\
		CFSS \cite{zhu2015face}   & 3,95  \\
		DSRN \cite{miao2018direct}   & 3.08  \\
		AAN \cite{yue2018attentional} & 2.99 \\
		\hline
		DeCaFA       & \textbf{2.10} \\
		\hline
	\end{tabular}
\end{table}

Finally, Table \ref{ablation3} shows a comparison of our method and state-of-the-art approaches on CelebA. As in \cite{Xiong2013,zhu2015face,miao2018direct,yue2018attentional} we report the average point-to-point error on the test partition, normalized by the distance between the two eye centers. Our approach is the best by a significant margin. Noteworthy, even though we use auxiliary data from 300W and WFLW, we do not use data from the \textit{val} partition of CelebA, contrary to \cite{miao2018direct,yue2018attentional}, thus there is significant room for improvement.

Overall, DeCaFA sets a new state-of-the-art on the three databases with several evaluation metrics. Also notice that it embraces few parameters ($\approx 10M$) compared to state-of-the-art approaches, and is also modular: at test time, DeCaFA allows to find a good trade-of between speed and accuracy (by evaluating only a fraction of the cascade), as well as to predict various numbers of landmarks.

\subsection{Weakly supervised learning}\label{weeakly}

\begin{figure*}[ht]
	\centering
	\includegraphics[width=\linewidth]{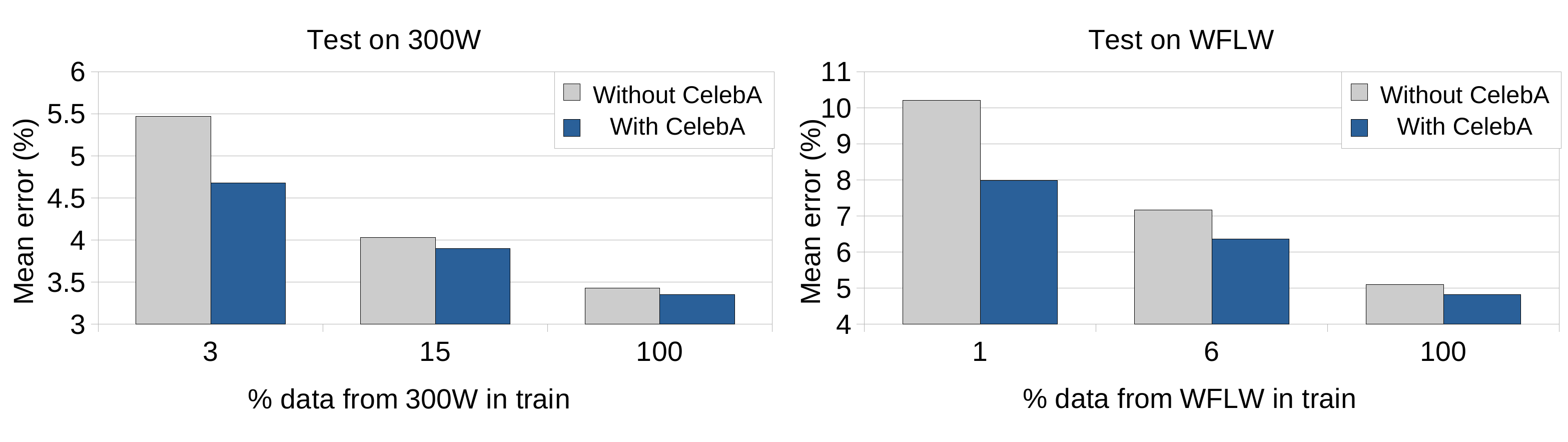}
	\caption{\% mean error comparison when training with small fraction of THE training set and coarsely annotated examples from CelebA.}
	\label{weakly}
\end{figure*}

In this Section, we study the capability of DeCaFA to learn with few examples annotated with $68$ and $98$-landmarks. To do so, we train DeCaFA using only a small, randomly sampled fraction of 300W (100 and 500 images, 3\% and 15\% of trainset) and WFLW (100 and 500 images,  1\% and 6\% of trainset) and the whole CelebA trainset, and report results on 300W and WFLW testsets on Figure \ref{weakly}.

Using coarsely annotated data from CelebA allows to substantially improve the landmark localization on both datasets, most notably when the number of training images is very low. For instance, DeCaFA trained with $3\%$ of 300W trainset and $1\%$ of WFLW trainset already outputs decent fine-grained landmark estimations, as it is better than CFSS \cite{zhu2015face} and DVLN (\cite{wu2017leveraging}, see Table \ref{ablation4}) on WFLW. DeCaFA trained with $15\%$ of 300W trainset and $6\%$ of WFLW trainset is on par with SAN on 300W (\cite{dong2018style}, see Table \ref{ablation2}), and is substantially better than DVLN on WFLW. This indicates that weakly supervised learning with examples from CelebA, that are coarsely annotated in terms of 5 landmarks, can significantly improve the prediction accuracy for the more precise tasks of predicting 68 and 98 landmarks. Thus, due to the chaining of multiple transfer layers, our DeCaFA architecture is well suited for this kind of weakly supervised learning and can be trained at a lower cost with coarsely annotated examples.

\subsection{Qualitative results}

\begin{figure*}[h!]
	\centering
	\includegraphics[width=\linewidth]{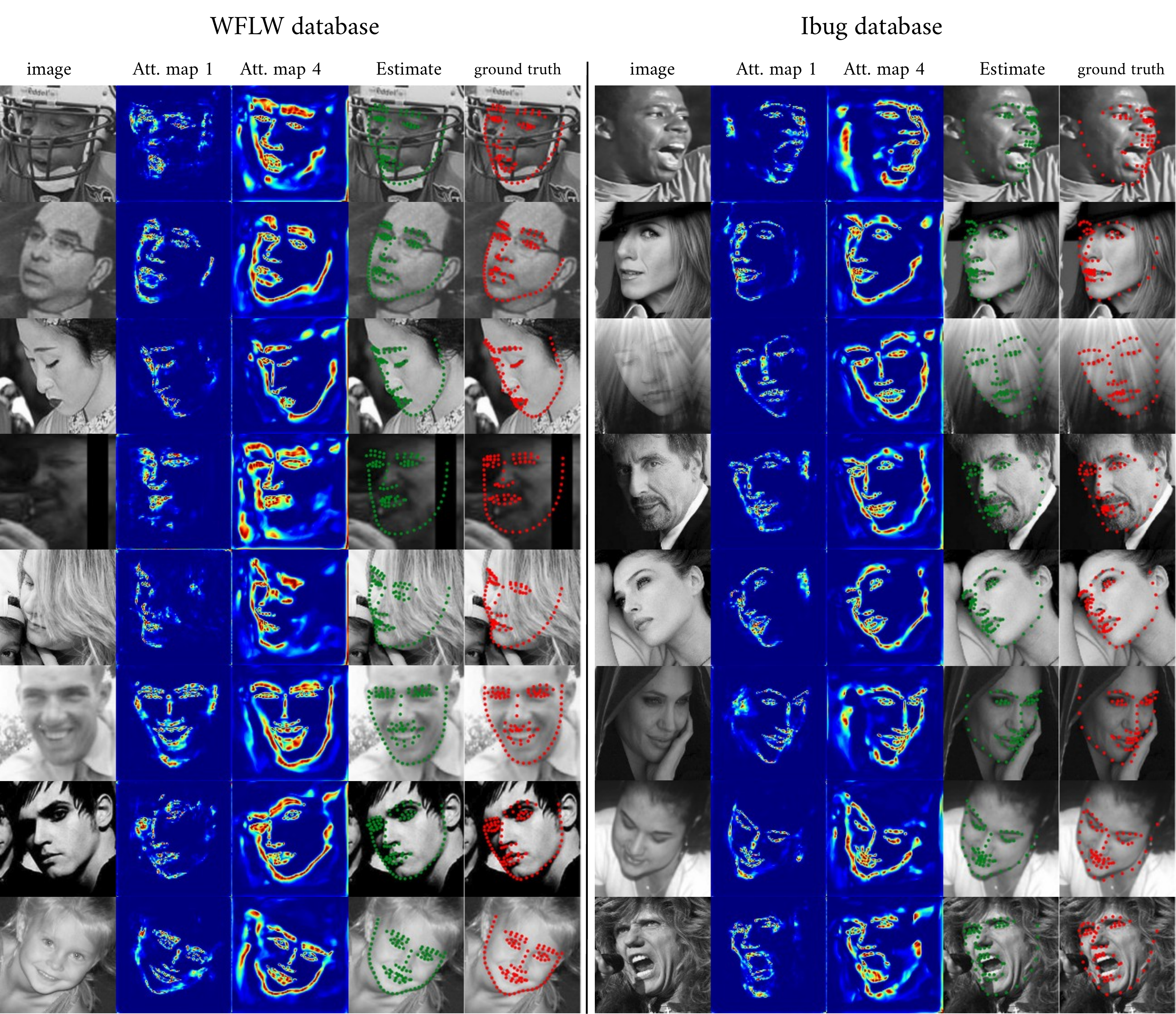}
	\caption{From left to right: images, attention maps outputted by stages 1 and 4, alignment results, and ground truth for images from 300W (I-bug, 68 landmarks) and WFLW (98 landmarks). Notice how the summed attention maps are iteratively refined, and how closely the predicted landmarks usually matches the ground truth, even under difficult illumination, non-frontal head poses, make-up, or occlusions.}
	\label{heatmaps}
\end{figure*}

Image \ref{heatmaps} shows vizualisations of aligned facial landmarks on WFLW, I-BUG and CelebA, as well as visualizations of attention maps after the cascade stages 1 and 4. Notice how these attention maps are coarse after stage 1 and get refined after stage 4, better highlighting the individual landmarks. Also notice that the predicted landmarks are close to the corresponding ground truth, even in the presence of rotations and occlusions (WFLW) or facial expressions (CelebA).

\section{Conclusion}

In this paper, we introduced DeCaFA for face alignment. DeCaFA unifies cascaded regression approaches and an end-to-end trainable deep methods. Its fully-convolutional U-net backbone ensures to keep full spatial resolution throughout the network, and the intermediate supervisions between the cascade stages with increasing weights ensures that the network learns cascaded alignment. Furthermore, by chaining multiple transfer layers to produce attention maps that correspond to multiple alignment tasks, DeCaFA can benefit from heterogeneous data. We empirically show that DeCaFA significantly outperforms state-of-the-art approaches on 300W, CelebA and WFLW databases. In addition, DeCaFA architecture is very modular and is suited for weakly supervised learning using coarsely annotated data with few landmarks.

Future work will consist in integrating other sources of data, or possibly other representations and tasks, such as head pose estimation, partial occlusion handling, as well as facial expressions, Action Unit and/or attributes (such as age or gender estimation) recognition within DeCaFA framework. Furthemore, we will study the application of DeCaFA to closely related fields, such as human pose estimation.

{\small
	\bibliographystyle{ieee}
	\bibliography{biblio}
}

\end{document}